\documentclass[pdflatex,sn-vancouver-num]{sn-jnl}% Springer Nature numbered Vancouver reference style

\usepackage{graphicx}%
\usepackage{amsmath,amssymb,amsfonts}%
\usepackage{booktabs}%
\usepackage{url}%
\let\orcidlogo\undefined
\usepackage{orcidlink}%
\usepackage{tikz}%
\usetikzlibrary{arrows.meta,positioning}%
\usepackage{longtable}%
\begin{document}

\title[Closed-loop Auto Research for molecular property prediction]{Closed-loop Auto Research for Molecular Property Prediction: Discovering and Certifying Generalizable Improvements}

\author*[1]{\fnm{Jingjie} \sur{Ning}\orcidlink{0009-0006-6854-0582}}\email{jening@cs.cmu.edu}

\author[1]{\fnm{Xiaochuan} \sur{Li}\orcidlink{0009-0004-9217-8072}}\email{xiaochu4@cs.cmu.edu}

\author[1]{\fnm{Ji} \sur{Zeng}}\email{jizeng@cs.cmu.edu}

\author[1]{\fnm{Chenyan} \sur{Xiong}\orcidlink{0000-0002-0392-4183}}\email{cx@cs.cmu.edu}

\author*[2]{\fnm{Guolin} \sur{Ke}\orcidlink{0000-0002-1227-7221}}\email{kegl@dp.tech}

\affil*[1]{\orgdiv{School of Computer Science}, \orgname{Carnegie Mellon University}, \orgaddress{\city{Pittsburgh}, \state{Pennsylvania}, \country{USA}}}

\affil[2]{\orgname{DP Technology}, \orgaddress{\city{Beijing}, \country{China}}}

%% BMC structured abstract.
\abstract{%
\textbf{Background.} Closed-loop Auto Research extends automated machine learning from fixed-dataset fitting to changing the research workflow, with language-model agents editing representations and model code and acquiring external evidence. Molecular property prediction spans many small endpoints. We ask whether this action space yields improvements generalizing beyond the validation signal selecting them.

\textbf{Results.} We isolate three Auto Research axes, features, models, and external evidence, under a file-level ablation lock attributing each gain to one axis over a strong baseline. Across 36 endpoints in three benchmark suites we score each selected configuration once on a held-out test whose labels the search never read. A routed pipeline taking each endpoint's best validation axis reaches positive held-out gains of 0.013, 0.011, and 0.042, the transferable axis differing by suite, data on TDC, model on Polaris, feature and model on MoleculeNet. The largest model-search gain falls from 0.041 on validation to 0.003 on test, while curated data reaches 0.022 but negative 0.019 on test, two non-transfer signatures. Curated external data raises held-out CYP2C9-substrate performance by 0.17 and half-life by 0.08, admitted through a contamination filter rejecting same-source files overlapping 64 to 89 percent of test structures, necessary but not sufficient for transfer. A matched-trial automated machine learning control did not reproduce the agent's code-level model intervention, reaching 0.006 against 0.042, and the pipeline stays competitive with an 84M-parameter pretrained 3D model on the shared training split.

\textbf{Conclusions.} The experiments stay within molecular property prediction, but separating discovery from held-out certification is a domain-agnostic lesson for any closed-loop system optimising a proxy for a held-out quantity.

\textbf{Scientific Contribution.} Closed-loop Auto Research, editing molecular representations, model code, and external evidence under a per-axis ablation lock, discovers improvements on a strong baseline that a matched-trial automated machine learning control does not reach. We pair this with held-out certification that freezes each validation-selected configuration, retrains it, and scores it once on unread test labels, separating transferable gains from two empirically distinct non-transfer signatures, selection variance and distribution shift. This separation, shown across three benchmarks, differentiates the work from prior workflows reporting only validation-selected performance.

}

\keywords{auto research, scientific agents, agentic experimentation, drug discovery, generalization, held-out evaluation, data leakage, molecular property prediction, ADMET prediction, external data augmentation}

\maketitle

% =====================================================================
\section{Introduction}\label{sec:intro}
Closed-loop Auto Research is a process in which language-model agents formulate hypotheses, make executable changes to a machine-learning pipeline, submit experiments to an evaluator they do not control, and update later hypotheses from measured outcomes \cite{ning2026autoresearch}. Its basic unit is a lineage of proposed, executed, and evaluated interventions. The point is broader than ordinary automated machine learning. Instead of only tuning a model on a fixed dataset, an Auto Research loop can change the representation, edit the estimator, acquire and vet new evidence, and revise the research programme as results accumulate. Unlike a loop that optimises and reports a single evaluator signal, the present study also acquires and audits external evidence and certifies each selected action on a held-out partition it never sees during the search.

Molecular property prediction is a high-friction setting for this kind of automation. Discovery workflows are fragmented across many heterogeneous endpoints, many endpoints are small, and domain users often need reliable task-specialized models without hand-building a new machine-learning recipe for each assay. Strong compact pipelines remain attractive because they are cheap to train, easy to deploy, and competitive with heavier pretrained models on many ADMET tasks. The promise of Auto Research is to turn such a compact baseline into a task-adapted model by deciding whether the next useful research action is a representation change, a model-code change, or new external evidence.

That expanded autonomy also changes the evaluation question. On a weak baseline an adaptive loop can almost always raise a validation score. On a strong, near-saturated baseline the scientifically meaningful question is whether the selected research actions generalize beyond the validation signal used to select them. A closed loop optimises that signal directly, so its reported gain is exposed to two distinct distortions. When the signal is a maximum over many trials on a small validation split, the selected configuration can reflect sampling noise rather than a reproducible effect. When the loop acquires external training data, the added records can be drawn from a different distribution than the evaluation set, or can re-import the benchmark's own measurements. Both mechanisms raise the validation score without improving genuine generalization. The appropriate unit of evaluation is therefore performance on a held-out test partition whose labels the search never reads. This reliability question and its resolution by held-out certification are methodological rather than chemical, even though we study them in molecular property prediction.

\begin{figure*}[t]
\centering
\includegraphics[width=\textwidth]{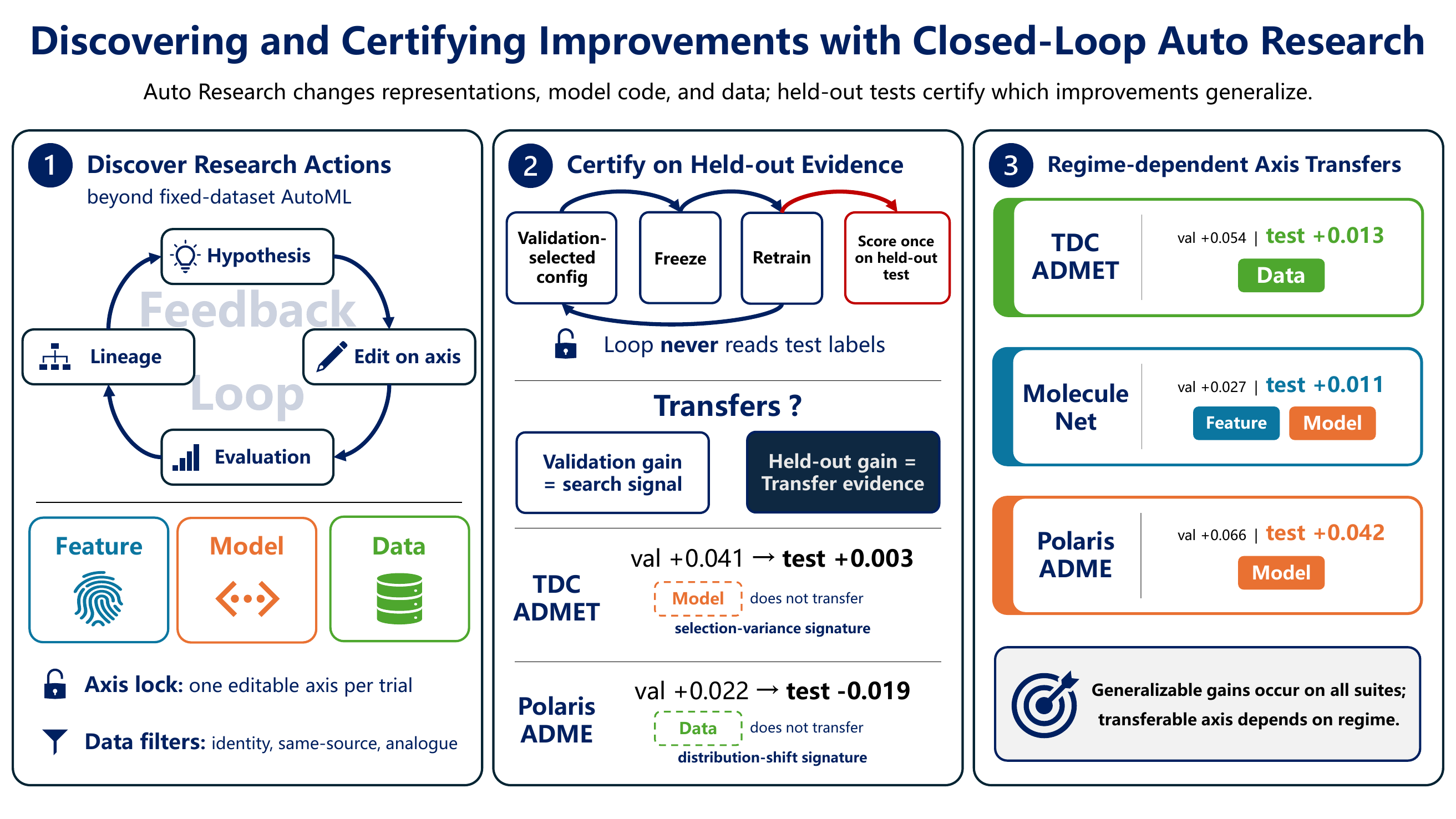}
\caption{Closed-loop Auto Research discovers research actions and certifies which generalize. (1)~Starting from a fixed, strong compact CPU baseline, the loop proposes, executes, and evaluates research actions that change the molecular representation, the model code and training procedure, or external evidence. A file-level ablation lock permits only one action class to change per trial, and external measurements pass identity, same-source, and near-analogue checks before training. (2)~The validation-selected configuration is frozen, retrained, and evaluated once on a held-out test partition whose labels are never read by the research loop. Paired validation and held-out gains identify both transfer and non-transfer signatures, illustrated by the TDC model action and the Polaris external-evidence action. (3)~Validation-routed configurations achieve positive held-out gains on all three suites, $+0.013$ on TDC ADMET, $+0.011$ on MoleculeNet, and $+0.042$ on Polaris ADME. The productive transferable action is regime-dependent, with external evidence leading on TDC, feature and model actions transferring on MoleculeNet, and model code leading on Polaris. Axis isolation attributes gains, while held-out evidence certifies which actions generalize.}
\label{fig:overview}
\end{figure*}

We study this problem for molecular property prediction, which offers heterogeneous endpoint regimes and permits repeated agent experiments without laboratory execution. We run \emph{axis-isolated Auto Research} over a strong MapLight-style baseline \cite{notwell2023admet} across three intervention axes. The feature axis changes molecular representation. The model axis changes the estimator, optimisation, and regularisation. The data axis changes the empirical evidence available for training through curated external measurements. A file-level ablation lock enforces a single active axis per trial, which attributes each measured change, and its generalization, to one research direction (Fig.~\ref{fig:overview}). We evaluate 36 endpoints across the TDC ADMET group \cite{huang2021tdc,huang2022artificial}, MoleculeNet \cite{wu2018moleculenet}, and the Biogen adme-fang dataset distributed through Polaris \cite{fang2023adme,polaris2024adme}.

After the searches complete we apply a held-out-test protocol. Each selected single-axis configuration is frozen, retrained on the internal training split, and scored once on the held-out test partition. The test labels are never read inside the loop, and the test structures are seen only by the evaluator-owned contamination filter described below. Auto Research produces transferable improvements on all three benchmarks, with the productive axis regime-dependent. Data leads on TDC, model search leads on Polaris, and both feature and model interventions transfer on MoleculeNet. Held-out evaluation then distinguishes these discoveries from validation-selected directions that do not transfer. The TDC model axis reaches $0.041$ on validation but $0.003$ on test, while Polaris data search reaches $0.022$ on validation and $-0.019$ on test, two empirically distinct non-transfer signatures. The same analysis identifies substantial endpoint-level discoveries: curated data raises held-out performance by $0.17$ on CYP2C9 substrate and $0.08$ on half-life.

A matched-trial AutoML control and a pretrained 3D comparator characterise these gains. A standard AutoML system searching the same model family on the same features at a matched trial count reaches a held-out improvement of $0.006$ on Polaris against $0.042$ for the agent, so the agent's code-level model intervention is not reproduced by off-the-shelf hyperparameter search. A pretrained 3D model with 84M parameters, fine-tuned on the same task-specific training split, is matched or exceeded by the CPU pipeline on most held-out endpoints, although the two methods do not share pretraining data or compute. The data axis is the source of the largest transferable gains on TDC, and acquiring and vetting external measurements lies outside the action space of a standard AutoML by construction.

\paragraph{Contributions.}
We make four contributions. First, axis-isolated closed-loop Auto Research turns a strong compact molecular-property baseline into task-adapted pipelines whose improvements transfer to a held-out test across all three benchmarks, reaching routed held-out gains of $+0.013$ on TDC, $+0.011$ on MoleculeNet, and $+0.042$ on Polaris, with the productive transferable axis depending on the task regime. Second, we define a held-out certification protocol that re-evaluates each validation-selected configuration once on an evaluator-isolated test, separating transferable discoveries from validation-only ones and documenting two empirically distinct non-transfer signatures, consistent with selection variance and with distribution shift. Third, we provide a leakage-safe external-evidence acquisition audit that makes every data decision reconstructable, including same-source rejections at 64 to 89 percent test overlap, for an action that lies outside fixed-dataset AutoML by construction. Fourth, we place the transferable gains against a matched-trial AutoML control and a pretrained 3D comparator that shares only the task-specific training split.

% =====================================================================
\section{Related work}\label{sec:related}
\textbf{Molecular property benchmarks and strong baselines.} MoleculeNet standardised molecular property prediction across quantum, physical chemistry, biophysical, and physiological tasks, along with common featurisations, splits, and learning models \cite{wu2018moleculenet}. TDC extended benchmark standardisation to therapeutics tasks, including the 22-endpoint ADMET group used here \cite{huang2021tdc,huang2022artificial}. These benchmarks motivated a large literature on learned molecular representations, including SMILES transformers such as ChemBERTa \cite{chithrananda2020chemberta}, self-supervised graph transformers such as GROVER \cite{rong2020grover}, contrastive graph pretraining such as MolCLR \cite{wang2022molclr}, and 3D representation learning such as Uni-Mol \cite{zhou2023unimol}. At the same time, strong fingerprint/descriptor baselines remain difficult to displace in ADMET settings. MapLight reports strong results from CatBoost \cite{prokhorenkova2018catboost} with combined ECFP \cite{rogers2010ecfp}, Avalon \cite{gedeck2006qsar}, ErG \cite{stiefl2006erg}, and RDKit descriptors across the TDC ADMET suite \cite{notwell2023admet}. We use a MapLight-style representation to test the remaining value of data, model, and feature changes.

\textbf{LLM agents for chemistry, drug discovery, and scientific work.} Chemistry-specific LLM agents have shown that tool use can extend language models beyond text-only chemical reasoning. ChemCrow integrated chemistry tools for tasks spanning molecular lookup, synthesis planning, and lab-connected workflows \cite{bran2024chemcrow}; Coscientist demonstrated autonomous planning and execution of chemical experiments through LLM tool use and laboratory APIs \cite{boiko2023coscientist}; the Virtual Lab used a multi-agent scientific team to design experimentally validated SARS-CoV-2 nanobodies \cite{swanson2025virtuallab}; and Co-Scientist introduced a multi-agent system for biomedical hypothesis generation with experimental validation \cite{gottweis2026coscientist}. In drug discovery, DrugAgent automates ML programming for representative drug-discovery tasks through planner/instructor agents \cite{liu2024drugagent}, while MolAgent provides an agentic framework for expert-level biomolecular property modelling with feature engineering, model selection, ensembling, validation, and 2D/3D representations \cite{gomeztamayo2025molagent}. These systems emphasize end-to-end task automation or high-fidelity automated modelling. Our evaluation separates data, model, and feature changes and measures their effects for each endpoint.

LLM4SD uses language models to synthesize literature knowledge and infer interpretable molecular features from data, with evaluation across molecular property tasks \cite{zheng2025llm4sd}. Its results establish that language-model derived chemical knowledge can improve prediction; we study a different level of the research process. The agent chooses and tests executable interventions across features, models, and external data, while an external evaluator records the resulting lineage. This separation allows us to ask which research action produces the gain and whether that action survives held-out certification.

\textbf{Agentic ML experimentation and auto research.} General-purpose benchmarks and systems have begun to evaluate LLM agents on complete machine-learning experiments. MLAgentBench studies language agents on iterative ML experimentation tasks \cite{huang2023mlagentbench}; MLE-bench evaluates agents on 75 Kaggle-style ML engineering competitions \cite{chan2024mlebench}; and AIDE explores code-space search for data-science workflows \cite{aide2025}. Broader automated-science systems such as The AI Scientist and AI Scientist-v2 extend this loop through ideation, experimentation, analysis, and manuscript production \cite{lu2026aiscientist,yamada2025aiscientistv2}. Our terminology of submitted trials, evaluator-owned outcomes, and shared lineage follows prior closed-loop auto-research work on compute-budgeted training recipes, where the loop optimises and reports a single evaluator signal under a fixed evaluator \cite{ning2026autoresearch}. We extend that loop in three directions specific to the reliability question. We re-evaluate each validation-selected configuration once on an evaluator-isolated held-out partition, we isolate the intervention axis responsible for each gain through a file-level ablation lock, and we add external-evidence acquisition with a contamination audit as a research action. These extensions let us ask not whether a loop can raise its own signal but whether the selected action transfers beyond it.

Recent benchmarks increasingly evaluate full-cycle research rather than isolated code generation. FIRE-Bench tests whether agents can rediscover scientific insights through executable experiments \cite{wang2026firebench}; ResearchGym places agents in repositories with strong human research baselines \cite{garikaparthi2026researchgym}; ResearchClawBench spans paper-grounded tasks across scientific domains \cite{xu2026researchclaw}; and SciAgentArena evaluates agents on interactive research problems across scales \cite{liu2026sciagentarena}. Together these studies expose a capability--reliability gap between occasionally strong research outcomes and consistent empirical support. Our work addresses a complementary question inside the experimental loop: when an agent adaptively discovers an improvement, does the selected intervention transfer to evaluator-isolated held-out evidence, and which class of research action produced that transfer? Axis isolation and paired validation/test evaluation make these questions measurable across feature, model, and data interventions.

Auto Research also differs from established automated machine learning. Bayesian optimisation and systems such as Auto-WEKA efficiently tune hyperparameters or select algorithms within a specified search space \cite{snoek2012bayesian,thornton2013autoweka}. Domain-specific systems such as Auto-ADMET further tailor ADMET pipelines to the chemical dataset at hand \cite{desa2025autoadmet}. That machinery can be part of an Auto Research system, but it normally operates within a fixed dataset and a predefined modelling search space. It does not by itself decide whether effort should go to representations, model code, new experimental evidence, or another scientific direction, and it does not audit newly acquired evidence for test contamination. Our study moves these research actions into the evaluated loop and asks which transfer to held-out evidence, aiming at attribution and certification across intervention classes rather than only optimisation within one class.

\textbf{Adaptive selection and held-out evidence.} The reliability question is not unique to language-model agents. Optimising a finite-sample model-selection criterion can overfit that criterion and bias subsequent performance estimates \cite{cawley2010overfitting}. More generally, adaptive data analysis studies how choices informed by previous measurements can compromise ordinary generalization guarantees, motivating restricted or reusable holdout mechanisms \cite{dwork2015adaptive,dwork2015reusable}. Closed-loop Auto Research is a new, high-capacity instance of this classical setting: hypotheses, code, and data can all change in response to the same validation signal. Reusable-holdout and differential-privacy mechanisms preserve validity by constraining access to the holdout; our setting instead keeps the research loop open-ended and evaluates frozen selections once after the search. We refer to these two strategies as constrain-during-search and certify-after-search. This connects autonomous research evaluation to adaptive empirical science while retaining the executable scope of modern research agents.

\textbf{Evaluation splits, leakage, and data sourcing.} Molecular ML evaluation is highly sensitive to how train/test splits represent deployment. MoleculeNet popularized scaffold splits as a more demanding alternative to random splits \cite{wu2018moleculenet}, and later work proposed still more realistic drug-discovery settings such as Lo-Hi hit-identification and lead-optimization benchmarks \cite{steshin2023lohi}. DataSAIL further formalizes split design to avoid information leakage in biological ML settings \cite{joeres2025datasail}. Across scientific machine learning, leakage has been documented as a broad source of irreproducible and overly optimistic claims \cite{kapoor2023leakage}. These concerns are amplified when autonomous systems can source external data and unintentionally import same-source molecules or near-duplicates, so we apply a harness-enforced leakage-safe filter to all external data (Section~\ref{sec:leak-protocol}).

% =====================================================================
\section{Experimental design}\label{sec:methods}

\subsection{Axis-isolated attribution and the reliability question}\label{sec:allocation}
Let $\mathcal{T}$ denote a set of prediction endpoints and let $\mathcal{A}=\{\mathrm{feature},\mathrm{model},\mathrm{data}\}$ denote the intervention axes. For axis $a$ the search retains complete pipeline snapshots, each scored on the internal validation split across all endpoints of a suite. We select the snapshot that maximises the mean validation improvement over those endpoints,
\begin{equation}
c_a=\arg\max_{c}\ \frac{1}{|\mathcal{T}|}\sum_{t\in\mathcal{T}} I^{\mathrm{val}}_t(c),
\end{equation}
where $I^{\mathrm{val}}_t(c)$ is the baseline-relative normalised improvement of snapshot $c$ on endpoint $t$. This single best-aggregate configuration $c_a$ is the unit the search reports for axis $a$. Its per-endpoint validation and held-out returns are
\begin{equation}
R^{\mathrm{val}}_{t,a}=I^{\mathrm{val}}_t(c_a),\qquad
R^{\mathrm{test}}_{t,a}=I^{\mathrm{test}}_t(c_a),
\end{equation}
where $R^{\mathrm{test}}$ re-evaluates the frozen $c_a$ once on the held-out test partition (Section~\ref{sec:heldout}). Both returns come from the same configuration, so the validation and held-out aggregates of an axis form a matched pair. The routed value selects, for each endpoint, the axis with the highest validation return above the threshold $\tau=0.005$ and reports that axis's held-out return.

The central object of this study is the relationship between the two returns. A research direction is reliable for an endpoint when its validation return is realised on the held-out test, and unreliable when the held-out return falls well below it. A closed loop reports $R^{\mathrm{val}}$ by construction and cannot, on its own, distinguish the two cases.

Axis isolation makes this comparison interpretable. Without it, a successful trial may combine a descriptor, an estimator, and an external dataset, so that neither the validation return nor its transfer can be attributed to a single research direction. Our design holds the evaluator, baseline, agent model, and trial contract fixed while changing only the editable intervention surface. The validation-to-test comparison is then made per axis under one experimental regime.

\subsection{Benchmark suites, splits, and metrics}\label{sec:env}
We evaluate 36 endpoint settings across three benchmark suites. The TDC ADMET Benchmark Group contributes 22 regression and binary classification endpoints. The official metrics comprise 5 MAE, 4 Spearman, 8 ROC-AUC, and 5 PR-AUC tasks. MoleculeNet contributes ten endpoints from FreeSolv, ESOL, BACE, HIV, Tox21, SIDER, and ClinTox \cite{wu2018moleculenet}. Regression is evaluated by RMSE and classification by ROC-AUC. The Biogen adme-fang dataset contributes human and rat liver microsomal clearance, MDR1-MDCK efflux, and aqueous solubility. These four log-transformed regression endpoints use Pearson correlation \cite{fang2023adme,polaris2024adme}.

The three suites differ in the provenance of their held-out test partition. TDC follows the benchmark's own scaffold partitions. For MoleculeNet we hold out a fixed 10 percent test partition per source file at a single seed, with a shared molecule partition across labels from the same file. For Polaris we hold out a scaffold-based 20 percent partition, which is stricter than the random split of \cite{fang2023adme}. Only the TDC partition is the published benchmark test split; the MoleculeNet and Polaris partitions are fixed outer held-out splits defined once and never re-drawn. For every suite the agent reward is computed on a scaffold-based 80:20 split carved from the outer training partition. The test labels are never read inside the agent loop, so every score the loop observes is computed on this internal validation split, and the test is scored only after the search under the protocol of Section~\ref{sec:heldout}.

For endpoint $t$, candidate metric $s_t$, and calibrated baseline metric $b_t$, the normalised improvement is
\begin{equation}
I_t = \begin{cases}
(s_t-b_t)/|b_t|, & \text{for higher-is-better metrics},\\
(b_t-s_t)/|b_t|, & \text{for lower-is-better metrics}.
\end{cases}
\end{equation}
The aggregate trial score is the unweighted mean of $I_t$ across the endpoints in a suite.

\subsection{Held-out-test protocol}\label{sec:heldout}
To measure generalization we re-evaluate the selected best-aggregate configuration of each axis after the search under a fixed protocol. We freeze the configuration, including its feature code, model code, and any admitted external data, retrain it from scratch on the internal training split, and predict once on the held-out test partition. The internal training and validation partitions are the same scaffold split (frac $0.20$, seed $42$) that produced the search reward, so the refitted validation score reproduces the reported one and serves as a consistency check while the held-out test is scored exactly once per configuration. For the data axis the external-data merge is replayed through the same leakage-safe filter (Section~\ref{sec:leak-protocol}), which reads the test structures but not the test labels and removes any external record that overlaps a test molecule before retraining. The held-out normalised improvement uses the same metric and direction as the search, computed against the baseline evaluated on the same test partition. This protocol reuses the search harness end to end and never modifies the selected configuration.

\subsection{Submitted-trial loop, specialists, and lineage}\label{sec:loop}

A trial follows four steps (Fig.~\ref{fig:overview}). An agent reads the current lineage, proposes a hypothesis, makes an executable change, and submits the resulting pipeline. The harness evaluates every endpoint in the active suite, assigns a status, and appends the score and feedback to the shared lineage. Later agents see this record and can retain, revise, or abandon the direction. The lineage therefore captures both successful interventions and research effort that failed to improve the evaluator outcome.

All specialists use the same language-model backend. The roles, evaluator contract, and ablation definitions remain fixed across suites, while endpoint descriptions and benchmark adapters change. Specialists are grouped by editable surface into feature, model, and external-evidence actions; a meta agent reads the lineage and coordinates direction. A \emph{file-level ablation lock} restores non-target pipeline files from a pristine copy before each subprocess, so a trial can edit only its assigned action class. The evaluator, metric computation, and leakage filter remain outside the editable pipeline (Section~\ref{sec:leak-protocol}). Implementation details are given in Appendix~\ref{sec:agents}.

\subsection{Search budgets and reporting units}\label{sec:budget}
The TDC study uses a budget of 100 trials per axis, with one additional feature trial; the two transfer studies use a target budget of 30 trials per axis. The reporting unit for every result in the main text is the single best-aggregate configuration $c_a$ of each axis defined in Section~\ref{sec:allocation}, scored as a matched validation and held-out-test pair. The routed value selects, per endpoint, the axis with the highest validation return among these configurations and reports its held-out score, so it reuses the same frozen configurations rather than a per-endpoint maximum taken over trials. The per-endpoint maxima over all retained snapshots are reported only as an exploratory validation analysis in Appendix~\ref{sec:valprofile} and never enter a configuration selection. Equal budgets allow a retrospective comparison of the axes and do not constitute an online policy that reallocates trials during search.

\subsection{Strong baseline}\label{sec:baseline}
The unedited pipeline follows MapLight \cite{notwell2023admet}: a fixed fingerprint/descriptor representation with CatBoost, MapLight target scaling for regression, and no per-task tuning. We recalibrate this baseline separately on each benchmark suite, which sets its aggregate improvement to $0$. On TDC, the MapLight-style pipeline improves 19 of 22 task metrics relative to a weaker fingerprint and XGBoost baseline, and it raises the small-magnitude Spearman score for half-life from $0.14$ to $0.40$. This strong starting point is part of the reliability question, since an intervention is useful only when it adds value beyond capabilities already present in the baseline. Full baseline composition and calibration details are provided in Appendix~\ref{sec:valleader}.

\subsection{Leakage-safe data augmentation}\label{sec:leak-protocol}
In \texttt{data\_only} mode the data agents add external labelled molecules via a controlled interface, and the ablation lock prevents simultaneous feature or model edits. Public molecular datasets frequently re-aggregate the same source assays, so an autonomous agent can unintentionally import test molecules or near-duplicates. The harness therefore applies a three-layer leakage-safe filter to every external file before it is merged, and the agent cannot bypass it.
\begin{enumerate}
\item \textbf{Standardised-identity de-duplication.} Each molecule is desalted, neutralised, and reduced to a standard InChIKey; rows matching any test, validation, or training molecule are removed. This catches salt, charge, and tautomer variants that exact-SMILES matching misses.
\item \textbf{Same-source rejection.} If more than $5\%$ of the endpoint's test molecules are present in the external file (by InChIKey skeleton), the entire file is rejected, since overlap at this scale indicates the benchmark's own or a sibling source.
\item \textbf{Near-analog filtering.} Molecules with ECFP4 Tanimoto $\geq 0.9$ to any test molecule are removed.
\end{enumerate}
Each merge is logged with source, overlap rate, and per-layer counts. The filter rejected same-source re-aggregations including human pharmacokinetic databases for half-life and VDss and a public mutagenicity set for AMES. Each source overlapped more than $60\%$ of the corresponding test set. The filter admitted independent curated data such as FDA-label pharmacokinetic values and literature CYP-substrate tables. The retained data-axis gains are therefore computed under enforced leakage control, and any source that bypasses same-source rejection is excluded from all comparisons.

% =====================================================================
\section{Held-out generalization}\label{sec:exp}
The baseline is a fixed MapLight-style pipeline (Section~\ref{sec:baseline}) whose validation scores match conventional CPU references across the three suites (Appendix~\ref{sec:valleader}). We evaluate generalization by freezing each selected configuration and scoring it once on the held-out test partition, with the same metric and normalisation as the search.

\begin{figure*}[tb]
\centering
\includegraphics[width=\textwidth]{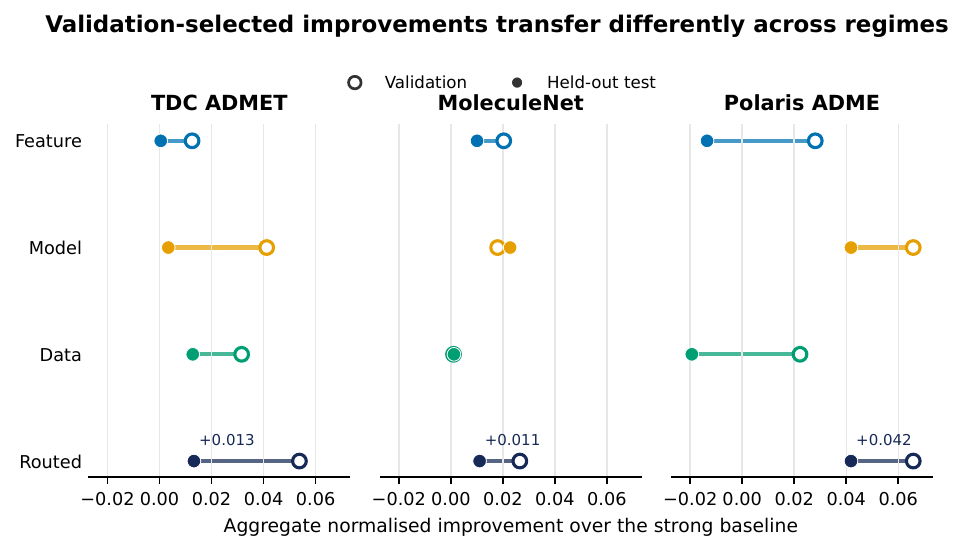}
\caption{Validation-selected improvements transfer differently across regimes. Open circles are aggregate validation gains and filled circles are held-out-test gains from the same frozen configuration, so each line measures transfer rather than a separately selected test result. Routed selects an axis per endpoint from validation and then reports that configuration on held-out data. Data is the leading transferable axis on TDC, while feature and model both transfer on MoleculeNet and model transfers on Polaris. Table~\ref{tab:heldout} lists the precise paired validation and held-out aggregates.}
\label{fig:transfer-map}
\end{figure*}

Figure~\ref{fig:transfer-map} makes the paired structure explicit. The same validation-selected configuration can retain, lose, or reverse its gain on the held-out partition, and the pattern changes across both axes and benchmark regimes.

Table~\ref{tab:heldout} reports, for each suite, the aggregate normalised improvement of every axis on the internal validation split and on the held-out test partition, with the routed value that selects the best axis per endpoint by validation and reads its held-out score. The validation column is the quantity the search optimised; the held-out column is the quantity that matters for a deployable model. Auto Research produces gains that transfer on all three suites. The transferable axis differs by suite, with data leading on TDC, model on Polaris, and feature and model both positive on MoleculeNet. The held-out column also identifies which validation gains do not transfer, and the size of that divergence depends on both the axis and the suite. On TDC the routed validation aggregate of $0.054$ falls to $0.013$ on the held-out test. On MoleculeNet the model axis grows, from $0.018$ on validation to $0.023$ on test. On Polaris the model axis retains most of its validation gain while the feature and data axes turn negative.

\begin{table*}[tb]
\caption{Validation against held-out-test generalization. Each cell is the aggregate normalised improvement over the strong baseline on the corresponding split. \emph{Routed} selects the best axis per endpoint by validation and reports its held-out score; selection never uses the test labels. Validation and held-out gains agree for the TDC data axis and for the MoleculeNet and Polaris model axes, and diverge for the TDC model axis and the Polaris data axis, in patterns consistent with selection variance and with distribution shift respectively. All columns are computed from the same frozen per-axis configurations. Figure~\ref{fig:transfer-map} shows the same pairs as validation-to-test transfer segments.}
\label{tab:heldout}
\centering
\small
\setlength{\tabcolsep}{5pt}
\begin{tabular*}{\textwidth}{@{\extracolsep{\fill}}lr rr rr rr rr@{}}
\toprule
& & \multicolumn{2}{c}{Feature} & \multicolumn{2}{c}{Model} & \multicolumn{2}{c}{Data} & \multicolumn{2}{c}{Routed} \\
\cmidrule(lr){3-4}\cmidrule(lr){5-6}\cmidrule(lr){7-8}\cmidrule(lr){9-10}
Suite & $n$ & val & test & val & test & val & test & val & test \\
\midrule
TDC ADMET         & 22 & $0.013$ & $+0.001$ & $0.041$ & $+0.003$ & $0.032$ & $+0.013$ & $0.054$ & $+0.013$ \\
MoleculeNet       & 10 & $0.020$ & $+0.010$ & $0.018$ & $+0.023$ & $0.001$ & $+0.001$ & $0.027$ & $+0.011$ \\
Polaris adme-fang & 4  & $0.028$ & $-0.014$ & $0.066$ & $+0.042$ & $0.022$ & $-0.019$ & $0.066$ & $+0.042$ \\
\bottomrule
\end{tabular*}
\end{table*}

\begin{figure*}[tb]
\centering
\includegraphics[width=\textwidth]{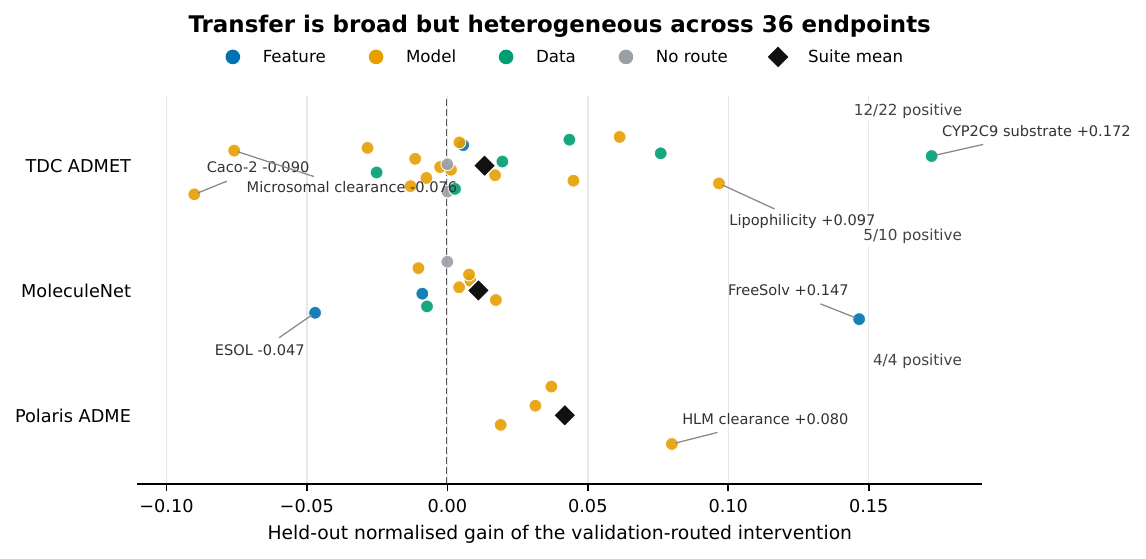}
\caption{Endpoint-level held-out gains of validation-routed interventions across the three benchmark suites. Each point is one endpoint and is coloured by the axis selected from validation; grey points have no axis above the routing threshold. Black diamonds mark suite means. Positive transfer occurs on 12 of 22 TDC endpoints, 5 of 10 MoleculeNet endpoints, and all 4 Polaris endpoints, with large gains on CYP2C9 substrate, FreeSolv, lipophilicity, and HLM clearance alongside annotated regressions on Caco-2, microsomal clearance, and ESOL. The complete endpoint table is reported in Appendix~\ref{sec:pertask}.}
\label{fig:endpoint-transfer}
\end{figure*}

The aggregate gains are not produced by a single endpoint. Figure~\ref{fig:endpoint-transfer} expands the routed result into all 36 held-out outcomes, showing transferable improvements across heterogeneous endpoint families together with the regressions that aggregate reporting would hide.

A single description organises Table~\ref{tab:heldout}. A validation gain transfers to the held-out test when the validation estimate tracks the held-out quantity, and fails to transfer when it does not. The data axis on TDC and the model axis on MoleculeNet and Polaris keep most of their improvement. The TDC model axis and the Polaris data axis lose almost all of theirs, and the loss takes two different forms that we examine next.

% =====================================================================
\section{What held-out certification reveals about transfer}\label{sec:failtransfer}

\subsection{Two non-transfer signatures}\label{sec:failure}
\begin{figure*}[tb]
\centering
\includegraphics[width=\textwidth]{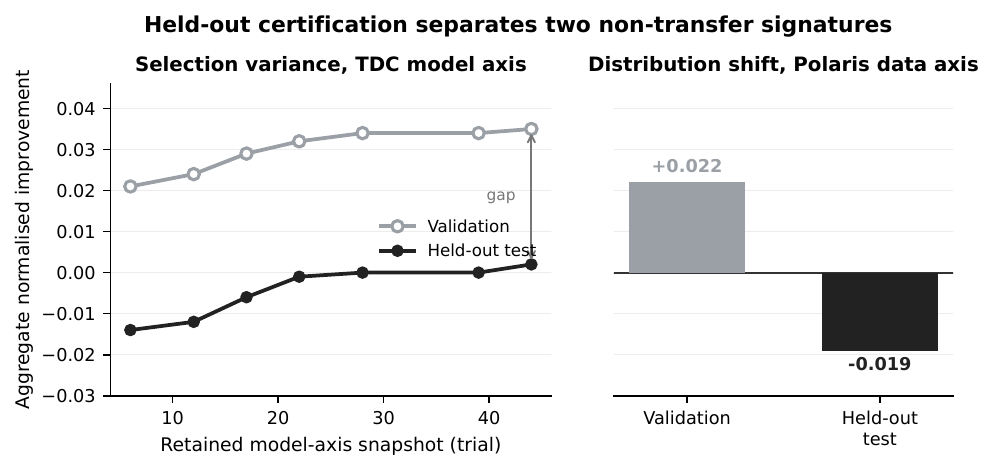}
\caption{Held-out certification separates two non-transfer signatures. On the TDC model axis (left) the validation aggregate climbs across retained snapshots while the held-out aggregate stays near zero, and the gap is present from the first snapshot, consistent with selection variance. On the Polaris data axis (right) the validation aggregate is positive while the held-out aggregate is negative, consistent with distribution shift. Validation is shown in grey and the held-out test in dark.}
\label{fig:signatures}
\end{figure*}

Held-out certification separates two empirically distinct non-transfer signatures (Figure~\ref{fig:signatures}).

The first pattern, which we call \emph{selection-variance non-transfer}, is consistent with selection variance, a known risk when model selection repeatedly adapts to a finite validation sample \cite{cawley2010overfitting,dwork2015reusable}. On TDC the model axis reaches a validation aggregate of $0.041$, the largest of any single axis, yet its held-out improvement is $0.003$. The validation gain is a maximum over one hundred submitted trials on a small validation split, so the selected configuration can capture split-specific structure that does not reappear on the test. As trials accumulate the model-axis validation aggregate rises steadily while its held-out aggregate, computed at each retained snapshot, stays near zero, and the gap is present from the first retained snapshot rather than opening late, consistent with selection pressure on a noisy validation estimate rather than late over-fitting of a single model.

The second pattern, which we call \emph{distribution-shift non-transfer}, is consistent with distribution shift. On Polaris the data axis reaches $0.022$ on validation and $-0.019$ on the held-out test. The external records the agent admitted raise the validation score but are drawn from assays that sit off-distribution relative to the Polaris test partition, and they lower test performance even though the contamination filter removed every test molecule from the merge. A higher validation score in this case reflects a change in the training distribution rather than a better estimate of the test quantity. This case also shows that contamination control alone does not guarantee transfer, since the same filter that secured the transferable TDC data gains passed the Polaris sources and the gain still did not generalize.

\subsection{Where data gains transfer, and the contamination filter}\label{sec:datatransfer}
The data axis on TDC is the clearest case in which a validation gain transfers. Its aggregate held-out improvement of $0.013$ is the largest single-axis test result on TDC, and it is concentrated where a compact independent assay changes coverage. Curated external data raises the held-out CYP2C9-substrate score by $0.17$ and half-life by $0.08$, with smaller positive transfer on CYP2D6 substrate and DILI. Not every source helps. CYP3A4-substrate data is slightly negative on the held-out test, so the data axis transfers on the endpoints where an independent source matches the assay and not uniformly.

A contamination filter screens every external source, and it is necessary for these gains without being sufficient. Across the three suites the same-source layer rejected three sources outright because they were the benchmark's own origin assays. A human pharmacokinetic database was rejected for half-life and for VDss at $88$ and $89$ percent test-structure overlap, and a mutagenicity benchmark was rejected for AMES at $64$ percent overlap. Admitting any of these would have produced a large but illusory validation gain. The eighteen sources that passed were independent, literature-cited assays such as FDA interaction guidance, the Obach 1999 clearance set, and DILIrank, and each was further scrubbed row by row for exact and near-analogue test matches before training. The filter controls contamination but not assay mismatch, which is why the Polaris data sources passed it and still failed to transfer (Section~\ref{sec:failure}). Every admission and rejection is reconstructable from the run log, which is the auditability property a drug-discovery setting requires.

\begin{figure*}[tb]
\centering
\includegraphics[width=\textwidth]{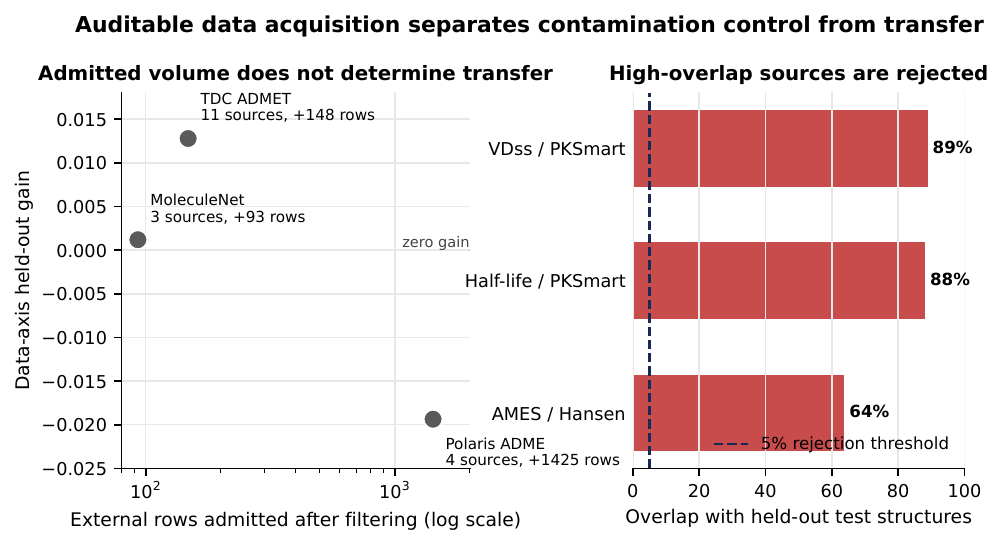}
\caption{Auditable external-data acquisition. Left, the 18 accepted sources add 1666 training rows across the three suites after identity and near-analogue filtering. The TDC data gain transfers, whereas the much larger Polaris merge is negative on held-out data, showing that contamination control is necessary but distributional match still determines transfer. Right, the evaluator-owned same-source rule rejects three proposed datasets whose held-out-structure overlap is 64--89\%, far above the 5\% rejection threshold.}
\label{fig:data-audit}
\end{figure*}

Figure~\ref{fig:data-audit} summarises the two roles of this audit. It blocks same-source contamination before training, and its accepted-source comparison shows that data volume alone does not determine held-out transfer.

What the loop changed is concrete on each axis. On the model axis it selected a per-task estimator among XGBoost, CatBoost, and LightGBM \cite{chen2016xgboost,prokhorenkova2018catboost,ke2017lightgbm}, matched capacity and regularisation to dataset size, and added a multi-seed ensemble that averages predictions across random seeds; these are code-level edits to the model file rather than a hyperparameter sweep over a fixed estimator. On the data axis it located named literature sources for individual endpoints, such as the Obach 1999 clearance set and DILIrank, and submitted each through the same-source and analogue filter before any row reached training. On the feature axis it added literature-grounded structural alerts, for example Kazius-style mutagenicity patterns and a lipophilic-basic-amine hERG pharmacophore. These interventions transfer selectively. The feature axis reaches $+0.010$ aggregate held-out improvement on MoleculeNet, led by a $+0.147$ gain on FreeSolv, while remaining close to zero in aggregate on TDC, where the 2563-dimensional baseline already captures much of the available representation signal. The complete per-axis discoveries are listed in Appendix~\ref{sec:valprofile}.

% =====================================================================
\section{External controls}\label{sec:controls}

\subsection{A matched-trial AutoML control on the model axis}\label{sec:automl}
The model axis transfers on Polaris, where it reaches a held-out improvement of $0.042$ and is the strongest axis, providing a clear case for testing whether standard hyperparameter search reproduces the gain. The control uses FLAML \cite{wang2019flaml} over the same model family, frozen MapLight features, scaffold split, and endpoint metrics, with the number of search trials matched to the agent. Table~\ref{tab:automl} reports the held-out result. The matched AutoML reaches an aggregate held-out improvement of $0.006$ against the agent's $0.042$, and supplying it with the same target transform the baseline and agent use does not close the gap. The agent's model-axis edits include a multi-seed ensemble that averages predictions across seeds, while the AutoML returns a single configuration selected on the same small validation split, a difference consistent with the variance pattern of Section~\ref{sec:failure} acting on the control. At the matched trial count, the control shows that the agent's transferable code-level intervention is not reducible to standard hyperparameter search.

\begin{table}[h]
\caption{Matched-trial AutoML control on the Polaris model axis, held-out test (Pearson). AutoML is FLAML over the same model family, features, split, and endpoint metrics at a matched trial count; the scaled column additionally applies the baseline target transform. The last row is the mean normalised improvement over the baseline.}
\label{tab:automl}
\centering
\small
\begin{tabular}{lrrrr}
\toprule
Endpoint & Baseline & Agent (model) & AutoML & AutoML (scaled) \\
\midrule
HLM clearance     & 0.682 & \textbf{0.737} & 0.732 & 0.732 \\
RLM clearance     & 0.744 & \textbf{0.758} & 0.741 & 0.731 \\
MDR1 permeability & 0.734 & \textbf{0.757} & 0.737 & 0.737 \\
Solubility        & 0.597 & \textbf{0.619} & 0.576 & 0.576 \\
\midrule
norm vs baseline  & --- & $+0.042$ & $+0.010$ & $+0.006$ \\
\bottomrule
\end{tabular}
\end{table}

\subsection{Comparison with a pretrained 3D model on held-out test}\label{sec:unimol}
We also evaluate Uni-Mol \cite{zhou2023unimol}, an 84M-parameter pretrained 3D model, on the held-out test partition of all three suites, fine-tuned on the same internal training split. The CPU pipeline is competitive. Across the 35 endpoints where Uni-Mol produces a valid test prediction, the strong baseline alone already exceeds Uni-Mol on 27, and the validation-selected pipeline on 25 (Table~\ref{tab:perendpoint}). Decomposed against the baseline, the held-out margin over Uni-Mol is carried mainly by the baseline, with the agent contributing a smaller increment on top. The exception is Polaris, where a 3D representation has a genuine advantage on RLM clearance and MDR1 permeability. There Uni-Mol exceeds the CPU baseline, and the agent's model axis narrows the gap without overturning it. The two methods share only the task-specific labelled training split, since Uni-Mol additionally uses its pretraining corpus and the agent uses curated external data, and their training procedures and compute differ. The comparison is therefore reported as evidence that a task-tuned CPU pipeline remains competitive with a heavy pretrained 3D model on the held-out test, rather than as a matched-budget result.

% =====================================================================
\section{Discussion}\label{sec:discussion}
Axis isolation makes the held-out analysis interpretable. Allowing simultaneous feature, model, and data edits would yield a stronger validation pipeline whose held-out behaviour could not be attributed to a research direction, and would mix a variance-prone axis with a bias-prone one. The file-level ablation lock lets each axis be evaluated for reliability on its own, which is how the two non-transfer signatures can be separated and named. The per-endpoint validation profile remains a useful diagnostic of where each kind of effort produced a measured signal, and its cross-suite ordering changes so that no axis is uniformly most productive. The held-out analysis supplies the necessary second step, since a productive validation direction is actionable only once its gain is shown to transfer.

The leakage-safe filter that governs acquired external data is part of held-out validity rather than implementation hygiene, and it is necessary rather than sufficient. It removes the failure that would most directly fake a data gain, the re-import of benchmark test molecules, but it does not control assay mismatch, so a source can pass the filter and still shift the training distribution away from the test, as the Polaris data axis shows. A transferable data gain therefore requires both contamination control and a source whose distribution matches the test, and only the first is enforced by the harness.

\section{Conclusions}\label{sec:conclusions}
This work studies closed-loop Auto Research, a mode of automation in which language-model agents do not merely tune the hyperparameters of a fixed pipeline but change the research workflow itself, rewriting the molecular representation, editing the predictive model code, and acquiring curated external evidence. The question we address is whether the improvements such a loop reports are real, that is, whether they survive on data the search never saw, or whether they are artefacts of repeatedly fitting the same validation signal. To answer it we pair two methods. The first is axis isolation, in which a file-level ablation lock confines each search run to exactly one kind of intervention, molecular features, predictive models, or external evidence, so that every measured gain over a strong compact MapLight-style baseline can be attributed to a single research direction. The second is held-out certification, in which each configuration selected on validation is frozen, retrained, and scored exactly once on a held-out test whose labels the search never read, which separates a genuine discovery from a validation artefact. We apply both across 36 endpoints in three molecular property prediction benchmark suites, the TDC ADMET collection, MoleculeNet, and the Biogen adme-fang dataset hosted on Polaris.

Across all 36 endpoints, a routed pipeline that takes each endpoint's best validation axis produced positive held-out gains on every suite, $+0.013$ on TDC, $+0.011$ on MoleculeNet, and $+0.042$ on Polaris, so held-out evaluation certifies the loop's discoveries rather than overturning them. The productive transferable direction changed with the benchmark regime, data curation on TDC, model intervention on Polaris, and feature and model intervention on MoleculeNet, so the useful research action is a property of the task environment and not a universal recipe. The largest endpoint-level effects came from leakage-controlled external data, which raised held-out CYP2C9-substrate performance by $0.17$ and half-life by $0.08$; these sources were admitted only through a contamination filter that rejected same-source files overlapping $64$ to $89$ percent of test structures, and contamination control proved necessary but not sufficient, because a source can pass the filter and still shift the training distribution. Two external controls place the loop against standard alternatives. A matched-trial automated machine learning control, FLAML at the same trial count on the Polaris model axis, reached only $0.006$ against the loop's $0.042$, because the loop's code-level model edits exceed the predefined hyperparameter space the control searched. External-evidence acquisition lies further outside fixed-dataset AutoML by construction, since the loop locates and vets new measurements rather than only tuning a model. On the shared training split the CPU pipeline stayed competitive with an 84M-parameter pretrained 3D model.

The same certification step exposed two cases in which a configuration that validation endorsed as an improvement did not carry over to the held-out test. We call these non-transfer signatures, and they are properties of the optimised feedback signal rather than of the agent. The first is selection-variance non-transfer, consistent with a reported gain that is a maximum over many submitted trials on a finite validation split; the TDC model axis reached the largest single-axis validation aggregate of $0.041$ yet only $0.003$ on the held-out test. The second is distribution-shift non-transfer, consistent with acquired data that moves the training distribution away from the test; the Polaris data axis reached $0.022$ on validation but $-0.019$ on the held-out test, even after every test molecule had been removed from the merge. Because both signatures are properties of any proxy that is optimised in place of a held-out quantity, they are systematic risks for any closed-loop research system, not quirks of this one.

Taken together, these results show that on molecular property prediction closed-loop Auto Research discovers improvements that reach beyond what fixed-dataset AutoML can change, and that held-out certification under axis isolation determines which of those discoveries transfer beyond the validation signal that selected them. The contribution extends past a molecular leaderboard without requiring the experimental domain to be broader than molecular property prediction. The underlying lesson is methodological and domain agnostic. Any closed-loop research system that proposes hypotheses, edits an executable workflow, and learns from a proxy measurement faces the same need to separate discovery from certification and the same two non-transfer signatures, since adaptive search can uncover real improvements while also amplifying noise or distribution mismatch. The chemistry-specific machinery, the identity, same-source, and near-analogue filters that govern molecular evidence acquisition, is what changes between domains; the certification logic does not. Axis isolation identifies where an improvement originates, and evaluator-isolated held-out evidence determines whether it is a transferable discovery.

% =====================================================================
\backmatter

\bmhead{Supplementary information}

Supplementary information is provided in the appendices, covering the search trajectories (Appendix~\ref{sec:traj}), per-endpoint results for the transfer suites, the specialist and tool configuration, the snapshot retention rule, the external-evidence action audit, the split configuration, and the validation intervention profile.

\bmhead{Acknowledgements}

Not applicable.

\section*{Declarations}

\bmhead{Funding}

This research received no specific grant from any funding agency in the public, commercial, or not-for-profit sectors.

\bmhead{Competing interests}

The authors declare no competing interests.

\bmhead{Ethics approval and consent to participate}

Not applicable.

\bmhead{Consent for publication}

Not applicable.

\bmhead{Availability of data and materials}

The harness traces, curated external-evidence files, configurations, and derived analysis tables that support the conclusions of this article are available in the repository \url{https://github.com/cxcscmu/Auto-Research-Molprop} under a Creative Commons Attribution 4.0 license. The benchmark datasets are public and cited in the text, namely the TDC ADMET Benchmark Group \cite{huang2021tdc,huang2022artificial}, MoleculeNet \cite{wu2018moleculenet}, and the Biogen adme-fang dataset on Polaris \cite{fang2023adme,polaris2024adme}. Third-party source datasets retain their original licenses.

\bmhead{Materials availability}

Not applicable.

\bmhead{Code availability}

The source code for the harness, task adapters, leakage-safe filter, and analysis scripts is available at \url{https://github.com/cxcscmu/Auto-Research-Molprop} under the Apache License 2.0.

\bmhead{Authors' contributions}

JN conceived the study, designed and implemented the Auto Research harness and the held-out certification protocol, ran the experiments, analysed the results, and drafted the manuscript. XL and JZ contributed to the implementation and the analysis. GK and CX supervised the work and revised the manuscript. All authors read and approved the final manuscript.

\bibliography{references}% common bib file

\clearpage
\appendix
\renewcommand{\thesection}{S\arabic{section}}
\renewcommand{\thefigure}{S\arabic{figure}}
\renewcommand{\thetable}{S\arabic{table}}
\renewcommand{\theHsection}{appendix.\thesection}
\renewcommand{\theHfigure}{appendix.\thefigure}
\renewcommand{\theHtable}{appendix.\thetable}
\setcounter{section}{0}
\setcounter{figure}{0}
\setcounter{table}{0}

\section{Search trajectories}\label{sec:traj}
\begin{figure}[h]
\centering
\includegraphics[width=0.78\textwidth]{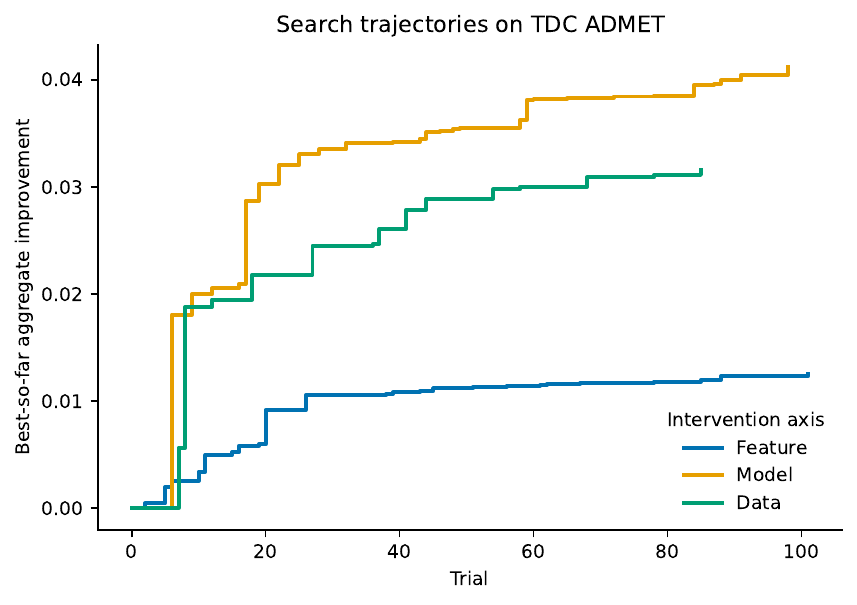}
\caption{Best-so-far aggregate normalised improvement against trial index for the three isolated axes on TDC ADMET.}
\label{fig:traj}
\end{figure}

Figure~\ref{fig:traj} reports the best-so-far aggregate normalised improvement on the TDC ADMET suite for each isolated axis across the budget of one hundred trials per axis, with one additional trial on the feature axis. The curve for each axis is the cumulative maximum of the aggregate score over its retained trials. Model search rises fastest and reaches the highest plateau, data search overtakes feature search near the tenth trial and plateaus below model, and feature search stays lowest throughout. The terminal values match the single-axis aggregates in the main text, namely $0.041$ for model, $0.032$ for data, and $0.013$ for feature.

Figure~\ref{fig:traj} tracks the validation aggregate. Table~\ref{tab:trajtest} re-evaluates the retained TDC model-axis snapshots on the held-out test partition in trial order. The validation aggregate rises from $0.021$ to $0.035$ across the sampled snapshots while the held-out aggregate stays near zero, and the gap is roughly constant from the first snapshot rather than opening with additional trials.

\begin{table}[h]
\caption{TDC model-axis validation-versus-held-out trajectory. Retained model-axis snapshots in trial order, each re-evaluated under the held-out protocol of Section~\ref{sec:heldout}. The validation aggregate climbs while the held-out aggregate stays near zero. This is the trajectory referenced in Section~\ref{sec:failure}.}
\label{tab:trajtest}
\centering
\small
\begin{tabular}{lrrr}
\toprule
Snapshot & VAL-norm & TEST-norm & Gap \\
\midrule
006 & $+0.021$ & $-0.014$ & $0.035$ \\
012 & $+0.024$ & $-0.012$ & $0.036$ \\
017 & $+0.029$ & $-0.006$ & $0.034$ \\
022 & $+0.032$ & $-0.001$ & $0.033$ \\
028 & $+0.034$ & $+0.000$ & $0.033$ \\
039 & $+0.034$ & $+0.000$ & $0.034$ \\
044 & $+0.035$ & $+0.002$ & $0.033$ \\
\bottomrule
\end{tabular}
\end{table}

\section{TDC validation intervention profile}\label{sec:valprofile}
On the validation split the single-axis aggregates are $0.041$ for model search, $0.032$ for data, and $0.013$ for feature. The per-endpoint validation profile (Table~\ref{tab:atlas}) selects model on 14 endpoints, data on 6, none on feature, and marks 2 flat; data gains concentrate on the small endpoints while model gains are broad. The interventions behind these numbers are concrete (Table~\ref{tab:discovery}). On the feature axis the agent assembled literature-grounded structural alerts such as Kazius-style mutagenicity patterns and a lipophilic-basic-amine hERG pharmacophore. On the model axis it selected a per-task estimator among XGBoost, CatBoost, and LightGBM, matched capacity and regularisation to dataset size, and added a multi-seed ensemble. On the data axis it located compact independent assays and named the compounds it added.

The cross-suite validation ordering changes. MoleculeNet places feature and model search close at $0.020$ and $0.018$ with data at $0.001$, while Polaris places model first at $0.066$ ahead of feature $0.028$ and data $0.022$. These validation profiles are exploratory diagnostics; the transferable directions are certified by the held-out results in the main text.

\begin{table*}[t]
\caption{Complete endpoint-level intervention profile on the validation split. Baseline is the absolute internal-validation score; Feature, Model, and Data give the largest normalised improvement per axis. Arrows give the metric direction; an axis is selected above the $0.005$ threshold.}\label{tab:atlas}
\centering
\scriptsize
\setlength{\tabcolsep}{3.3pt}
\begin{tabular}{lrrrrrrc}
\toprule
Endpoint & $n_{\mathrm{train}}$ & Metric & Baseline & Feature & Model & Data & Selected \\
\midrule
Caco-2 Wang            & 583  & MAE$\downarrow$      & 0.369 & +0.041 & \textbf{+0.131} & +0.059 & Model \\
HIA Hou                 & 369  & AUROC$\uparrow$       & 0.980 & +0.002 & +0.001 & +0.000 & Flat \\
P-gp Broccatelli       & 778  & AUROC$\uparrow$       & 0.919 & +0.002 & +0.004 & \textbf{+0.005} & Data \\
Bioavailability Ma     & 410  & AUROC$\uparrow$       & 0.727 & +0.016 & \textbf{+0.086} & +0.044 & Model \\
Lipophilicity AZ       & 2688 & MAE$\downarrow$       & 0.491 & +0.003 & \textbf{+0.069} & +0.000 & Model \\
Solubility AqSolDB     & 5399 & MAE$\downarrow$       & 0.649 & +0.000 & \textbf{+0.037} & +0.000 & Model \\
BBB Martins            & 1250 & AUROC$\uparrow$       & 0.885 & +0.009 & \textbf{+0.024} & +0.004 & Model \\
PPBR AZ                & 1785 & MAE$\downarrow$       & 9.965 & +0.002 & \textbf{+0.024} & +0.000 & Model \\
VDss Lombardo          & 719  & Spearman$\uparrow$    & 0.716 & +0.024 & +0.024 & \textbf{+0.030} & Data \\
CYP2D6 Veith           & 8404 & AUPRC$\uparrow$       & 0.676 & +0.008 & \textbf{+0.031} & +0.000 & Model \\
CYP3A4 Veith           & 7889 & AUPRC$\uparrow$       & 0.858 & +0.000 & \textbf{+0.011} & +0.000 & Model \\
CYP2C9 Veith           & 7737 & AUPRC$\uparrow$       & 0.795 & +0.003 & +0.002 & +0.000 & Flat \\
CYP2D6 substrate       & 426  & AUPRC$\uparrow$       & 0.585 & +0.037 & +0.052 & \textbf{+0.141} & Data \\
CYP3A4 substrate       & 426  & AUROC$\uparrow$       & 0.629 & +0.021 & \textbf{+0.097} & +0.057 & Model \\
CYP2C9 substrate       & 428  & AUPRC$\uparrow$       & 0.408 & +0.053 & +0.073 & \textbf{+0.225} & Data \\
Half-life Obach        & 426  & Spearman$\uparrow$    & 0.396 & +0.000 & +0.036 & \textbf{+0.063} & Data \\
Microsomal clearance   & 697  & Spearman$\uparrow$    & 0.640 & +0.038 & \textbf{+0.090} & +0.051 & Model \\
Hepatocyte clearance   & 776  & Spearman$\uparrow$    & 0.558 & +0.000 & \textbf{+0.078} & +0.000 & Model \\
hERG                    & 419  & AUROC$\uparrow$       & 0.849 & +0.006 & \textbf{+0.007} & +0.000 & Model \\
AMES                    & 3963 & AUROC$\uparrow$       & 0.794 & +0.001 & \textbf{+0.022} & +0.000 & Model \\
DILI                    & 304  & AUROC$\uparrow$       & 0.862 & +0.014 & +0.033 & \textbf{+0.036} & Data \\
LD50 Zhu                & 3973 & MAE$\downarrow$       & 0.464 & +0.000 & \textbf{+0.019} & +0.000 & Model \\
\bottomrule
\end{tabular}
\end{table*}

\begin{table}[tb]
\caption{Representative interventions discovered on each axis, with the observed validation effect over the strong baseline.}
\label{tab:discovery}
\centering
\small
\begin{tabular}{@{}l p{0.46\textwidth} p{0.20\textwidth}@{}}
\toprule
Axis & Representative discovery & Observed effect \\
\midrule
Feature    & Mutagenicity alerts (aziridine, Michael acceptor, nitrogen mustard); hERG benzhydryl and N-arylpiperazine; DILI thiazolidinedione & Small; signal saturated by the baseline representation \\
Model      & Per-task estimator family; capacity and regularization matched to dataset size; multi-seed ensemble & Largest aggregate gain \\
Data       & Curated independent assays (DILIrank, Obach 1999, FDA labels) with named compounds & Large on small endpoints \\
\bottomrule
\end{tabular}
\end{table}

\section{Per-endpoint held-out generalization}\label{sec:pertask}
Table~\ref{tab:perendpoint} gives, for every endpoint, the baseline held-out-test score, the axis selected by validation (best of feature F, model M, or data D above the $0.005$ threshold, or none), the validation gain of that selected configuration, its held-out test gain, and the held-out normalised improvement of Uni-Mol against the same baseline. This is the per-endpoint evidence behind the suite-level aggregates of Table~\ref{tab:heldout}. The selected axis transfers on the data-rich small endpoints (CYP2C9 and CYP2D6 substrate, half-life, DILI) and on the MoleculeNet and Polaris model endpoints, and fails on the TDC model endpoints where the validation gain does not reappear on the test. Against Uni-Mol the validation-selected pipeline wins on 25 of the 35 endpoints with a valid Uni-Mol prediction and the baseline alone on 27, so the margin over the pretrained model is carried mainly by the strong baseline.

{\small
\begin{longtable}{llrcrrr}
\caption{Per-endpoint paired validation and held-out-test generalization. Baseline is the absolute held-out-test score. Sel.\ is the validation-selected axis (F/M/D, or --- if none clears $0.005$). Val gain and Test gain are the normalised improvement of that selected configuration on validation and on the held-out test. Uni-Mol is its held-out normalised improvement over the same baseline (-- where 3D conformer generation fails).}
\label{tab:perendpoint}\\
\toprule
Endpoint & Metric & Baseline & Sel. & Val gain & Test gain & Uni-Mol \\
\midrule
\endfirsthead
\toprule
Endpoint & Metric & Baseline & Sel. & Val gain & Test gain & Uni-Mol \\
\midrule
\endhead
\bottomrule
\endfoot
\multicolumn{7}{l}{\emph{TDC ADMET}}\\
Caco-2 Wang & mae & 0.292 & M & +0.131 & $-0.090$ & $-0.319$ \\
HIA Hou & roc-auc & 0.975 & --- & +0.000 & +0.000 & $-0.017$ \\
P-gp Broccatelli & roc-auc & 0.926 & D & +0.005 & +0.003 & $-0.053$ \\
Bioavailability Ma & roc-auc & 0.736 & M & +0.086 & $-0.013$ & $-0.119$ \\
Lipophilicity AZ & mae & 0.558 & M & +0.069 & +0.097 & +0.153 \\
Solubility AqSolDB & mae & 0.794 & M & +0.037 & +0.045 & -- \\
BBB Martins & roc-auc & 0.919 & M & +0.024 & $-0.007$ & $-0.082$ \\
PPBR AZ & mae & 7.893 & M & +0.024 & +0.017 & $-0.031$ \\
VDss Lombardo & spearman & 0.712 & D & +0.030 & $-0.025$ & $-0.123$ \\
CYP2D6 Veith & pr-auc & 0.715 & M & +0.031 & +0.001 & $-0.071$ \\
CYP3A4 Veith & pr-auc & 0.876 & M & +0.008 & $-0.003$ & $-0.046$ \\
CYP2C9 Veith & pr-auc & 0.776 & --- & +0.000 & +0.000 & $-0.030$ \\
CYP2D6 substrate & pr-auc & 0.708 & D & +0.141 & +0.020 & $-0.196$ \\
CYP3A4 substrate & roc-auc & 0.652 & M & +0.083 & $-0.011$ & +0.048 \\
CYP2C9 substrate & pr-auc & 0.373 & D & +0.225 & +0.172 & $-0.125$ \\
Half-life Obach & spearman & 0.519 & D & +0.063 & +0.076 & $-0.135$ \\
Microsomal clearance & spearman & 0.625 & M & +0.078 & $-0.076$ & $-0.051$ \\
Hepatocyte clearance & spearman & 0.452 & M & +0.066 & $-0.028$ & $-0.027$ \\
hERG & roc-auc & 0.886 & F & +0.006 & +0.006 & $-0.126$ \\
AMES & roc-auc & 0.844 & M & +0.022 & +0.004 & $-0.044$ \\
DILI & roc-auc & 0.840 & D & +0.036 & +0.043 & +0.050 \\
LD50 Zhu & mae & 0.610 & M & +0.019 & +0.061 & +0.069 \\
\multicolumn{7}{l}{\emph{MoleculeNet}}\\
FreeSolv & rmse & 0.909 & F & +0.099 & +0.147 & $-0.763$ \\
ESOL & rmse & 0.717 & F & +0.054 & $-0.047$ & +0.014 \\
BACE & roc-auc & 0.842 & D & +0.006 & $-0.007$ & $-0.012$ \\
HIV & roc-auc & 0.750 & M & +0.021 & +0.017 & +0.037 \\
Tox21 NR-AR & roc-auc & 0.720 & F & +0.016 & $-0.009$ & $-0.103$ \\
Tox21 SR-MMP & roc-auc & 0.926 & M & +0.010 & +0.004 & $-0.042$ \\
Tox21 SR-p53 & roc-auc & 0.815 & M & +0.009 & +0.008 & $-0.045$ \\
SIDER hepatobiliary & roc-auc & 0.748 & M & +0.013 & +0.008 & $-0.131$ \\
SIDER reproductive & roc-auc & 0.756 & M & +0.037 & $-0.010$ & $-0.136$ \\
ClinTox & roc-auc & 0.951 & --- & +0.000 & +0.000 & $-0.162$ \\
\multicolumn{7}{l}{\emph{Polaris ADME}}\\
HLM clearance & pearson & 0.682 & M & +0.079 & +0.080 & $-0.009$ \\
RLM clearance & pearson & 0.744 & M & +0.059 & +0.019 & +0.039 \\
MDR1 permeability & pearson & 0.734 & M & +0.033 & +0.031 & +0.055 \\
Solubility & pearson & 0.597 & M & +0.093 & +0.037 & $-0.003$ \\
\bottomrule
\end{longtable}
}

\section{Baseline calibration}\label{sec:valleader}
The fixed MapLight-style baseline uses 1024 Morgan counts \cite{rogers2010ecfp}, 1024 Avalon counts \cite{gedeck2006qsar}, 315 ErG features \cite{stiefl2006erg}, and 200 RDKit descriptors \cite{rdkit}, followed by CatBoost with the MapLight regression target transform. Table~\ref{tab:baseline} calibrates this fixed baseline against published CPU references. Published values come from different train/test protocols, so the table is contextual rather than a like-for-like leaderboard.

\begin{table}[tb]
\caption{Baseline calibration against published CPU references. Baseline is the internal-validation score of the fixed MapLight pipeline; published values are test scores, so the comparison is contextual rather than like-for-like. $\dagger$ Tox21 12-task and $\ddagger$ SIDER 27-task averages of \cite{wu2018moleculenet}; ESOL and FreeSolv use a random split. Polaris references are from \cite{fang2023adme}.}
\label{tab:baseline}
\centering
\small
\begin{tabular}{llrl}
\toprule
Suite / endpoint & Metric & Baseline & Published CPU reference \\
\midrule
\multicolumn{4}{l}{\emph{MoleculeNet} \cite{wu2018moleculenet}} \\
ESOL                & RMSE$\downarrow$    & 0.69  & 0.99 (RF, XGBoost) \\
FreeSolv            & RMSE$\downarrow$    & 1.96  & 1.74 (XGBoost) \\
BACE                & ROC-AUC$\uparrow$   & 0.885 & 0.867 (RF) \\
HIV                 & ROC-AUC$\uparrow$   & 0.792 & 0.792 (kernel SVM) \\
Tox21 NR-AR         & ROC-AUC$\uparrow$   & 0.831 & 0.822$^\dagger$ (kernel SVM) \\
Tox21 SR-MMP        & ROC-AUC$\uparrow$   & 0.891 & 0.822$^\dagger$ \\
Tox21 SR-p53        & ROC-AUC$\uparrow$   & 0.844 & 0.822$^\dagger$ \\
SIDER hepatobiliary & ROC-AUC$\uparrow$   & 0.712 & 0.684$^\ddagger$ (RF) \\
SIDER reproductive  & ROC-AUC$\uparrow$   & 0.691 & 0.684$^\ddagger$ \\
ClinTox             & ROC-AUC$\uparrow$   & 0.966 & n/a \\
\midrule
\multicolumn{4}{l}{\emph{Polaris adme-fang} \cite{fang2023adme}} \\
HLM clearance       & Pearson$\uparrow$   & 0.678 & RF, GBDT \\
RLM clearance       & Pearson$\uparrow$   & 0.699 & RF, GBDT \\
MDR1-MDCK           & Pearson$\uparrow$   & 0.709 & RF, GBDT \\
Solubility          & Pearson$\uparrow$   & 0.540 & RF, GBDT \\
\bottomrule
\end{tabular}
\end{table}

\section{Specialists and tool interface}\label{sec:agents}
The eight specialists share one Claude Sonnet 4.6 model and differ only in their editable surface and prompt. The feature group (\texttt{fphs}, \texttt{fsub}, \texttt{lit}) edits the molecular representation through physicochemical descriptors, substructure and SMARTS alerts, and literature-grounded features. The model group (\texttt{modl}, \texttt{calib}) edits the estimator family, its hyperparameters, and probability calibration. The data group (\texttt{data}, \texttt{daugm}) edits data handling and acquires external labelled molecules through a single controlled tool, \texttt{write\_external\_data}, which writes one comma-separated file per endpoint. The \texttt{meta} agent reads the lineage and coordinates direction. Before every trial subprocess a file-level ablation lock restores all files outside the active axis from a pristine copy, so a feature trial cannot alter the model and a data trial can change only the external-data output. The evaluator, the metric computation, and the leakage filter sit outside the editable pipeline and are never modified by an agent. Environment stripping, benchmark-library removal from the agent environment, and a mount-namespace barrier provide additional isolation. The three axes were searched sequentially, and within each axis the trials run one at a time so that every agent reads the updated lineage before proposing the next change.

\section{Snapshot retention}\label{sec:retention}
Each submitted pipeline is evaluated on every endpoint and assigned a status. A trial that improves the lineage receives a stored pipeline snapshot together with its per-endpoint scores; the remaining trials are recorded in the lineage without a snapshot. The trajectories in Appendix~\ref{sec:traj} and the per-endpoint maxima in the validation profile are computed over the retained snapshots, of which there are twenty-four for the feature axis, thirty for the model axis, and fourteen for the data axis on TDC.

\section{External-evidence action audit}\label{sec:audit}
The leakage-safe filter records a verdict for every external-evidence action before any proposed row is merged. On TDC the agents repeatedly proposed large public pharmacokinetic and toxicity databases that re-aggregate benchmark sources, and the evaluator rejected them as same-source. Examples include human half-life and volume-of-distribution tables whose InChIKey-skeleton overlap with the test set exceeded eighty-five percent, and a mutagenicity set whose overlap exceeded sixty percent. Independent curated sources were admitted, including a small drug-drug-interaction CYP-substrate table that produced the largest single data gain on CYP2C9 substrate. The three filter layers are standardised-identity de-duplication, whole-file same-source rejection above five percent test overlap, and near-analogue removal at ECFP4 Tanimoto $\geq 0.9$.

\begingroup
\footnotesize
\noindent\textbf{Table S7. Representative external-evidence decisions.} The table reports the agent's intended evidence action, the evaluator-owned audit decision, and the held-out consequence when the action was admitted. The full per-file log is stored in the lineage.\par\smallskip

\setlength{\tabcolsep}{2pt}
\begin{tabular}{@{}p{0.18\textwidth}p{0.44\textwidth}p{0.30\textwidth}@{}}
\toprule
Target & Evidence action and audit decision & Consequence \\
\midrule
Half-life & Public human PK aggregate; rejected as same-source ($\approx$88\% overlap) & Not merged; blocks benchmark re-aggregation \\
VDss & Public human PK aggregate; rejected as same-source ($\approx$89\% overlap) & Not merged; blocks sibling-source leakage \\
AMES & Public mutagenicity set; rejected as same-source ($\approx$64\% overlap) & Not merged; public source can be a benchmark sibling \\
CYP2C9 substrate & FDA drug-interaction substrate table; accepted after identity/analogue scrub & $+0.172$ held-out gain; matched evidence transfers \\
Half-life & FDA clinical-pharmacology records; accepted after row-level scrub & $+0.076$ held-out gain; small assays change coverage \\
Polaris ADME & ChEMBL/ESOL-derived ADME records; accepted after row-level scrub & $-0.019$ aggregate held-out gain; audit does not ensure distributional match \\
\bottomrule
\end{tabular}
\par \endgroup

\section{Split configuration}\label{sec:splits}
For TDC the outer train and test partition is the one distributed with the benchmark group. For MoleculeNet each dataset uses its recommended split, scaffold for ESOL, BACE, and HIV and random for FreeSolv, Tox21, SIDER, and ClinTox, with a single shared molecule partition across labels from the same source file. For Polaris the outer split is a scaffold-based 80:20 partition. In every suite the agent reward is computed on a further scaffold-based 80:20 split carved from the outer training partition with seed 42, and the outer test labels are never exposed to the agent loop.

\end{document}